\documentclass{llncs}
\usepackage{graphicx}
\begin{document}

\frontmatter          
\pagestyle{headings}  
\addtocmark{The NUbots Qualification material for RoboCup 2014} 
\mainmatter              
\title{The NUbots Team Description Paper 2014}
\titlerunning{The NUbots Team Description for 2014}  

\author{Josiah Walker
		\and Trent Houliston
		\and Brendan Annable
		\and Alex Biddulph
		\and Andrew Dabson
		\and Jake Fountain
		\and Taylor Johnson
		\and Jordan Johnson
		\and Mitchell Metcalfe
		\and Anita Sugo
		\and Stephan K. Chalup
		\and Robert A.R. King
		\and Alexandre Mendes
		\and Peter Turner}
\authorrunning{Walker et al.}   
%
\tocauthor{J. Walker,
T. Houliston,
B. Annable,
A. Biddulph,
A. Dabson,
J. Fountain,
T. Johnson,
J. Johnson,
M. Metcalfe,
A. Sugo,
S. Chalup,
R.A.R. King,
A. Mendes,
P. Turner}

\institute{Newcastle Robotics Laboratory\\ School of Electrical Engineering \& Computer Science\\
Faculty of Engineering and Built Environment\\
The University of Newcastle, Callaghan 2308, Australia\\
Contact: \email{stephan.chalup@newcastle.edu.au}\\
Homepage: \texttt{http://robots.newcastle.edu.au}}

\maketitle              

\begin{abstract}
The NUbots team, from The University of Newcastle, Australia, has had a strong record of success in the RoboCup Standard Platform League since first entering in 2002. The team has also competed within the RoboCup Humanoid Kid-Size League since 2012. The 2014 team brings a renewed focus on software architecture, modularity, and the ability to easily share code. This paper summarizes the history of the NUbots team, describes the roles and research of the team members, gives an overview of the NUbots' robots and software system, and addresses relevant research projects within the the Newcastle Robotics Laboratory.
\end{abstract}

\section{Introduction}
The NUbots team, from the University of Newcastle, Australia, competed in the Four Legged League from 2002-2007, within the Standard Platform League from 2008-2011 and subsequently within the Kid-Size Humanoid league since 2012. The NUbots have had a strong record of successes, twice achieving a first place; in 2006 in Bremen, Germany, and, again in 2008 as part of the NUManoid team in Suzhou, China.

The central goal of the NUbots is to be a high performance competitive robot soccer team at RoboCup. The vision of the research projects associated with the NUbots team is to develop and program robots that can support humans not only for routine, challenging, or dangerous tasks, but also to improve quality of life through personal assistance, companionship, and coaching.  Our mission is to contribute to a responsible development and application of robotics. Some of our projects therefore emphasise anthropocentric and biocybernetic
aspects in robotics research~\cite{ChalupOstwald2009}. This includes new aspects of human robot interaction and perception. The Newcastle Robotics Laboratory hosts several postgraduate and undergraduate research projects that are associated with the NUbots.


\section{Commitment to RoboCup 2014}
The Nubots commit to participation at RoboCup 2014 upon successful qualification. We also commit to provision of a person, with sufficient knowledge of the rules, available as referee during the competition.

\section{History of the NUbots' participation at RoboCup}

The NUbots team was founded in 2002 and participated for the first time at RoboCup in Fukuoka in the Sony Four-Legged League (3rd place). Since then the team has  a strong history of competition and success in the RoboCup SPL/Four-Legged League, obtaining many top three placements and winning the title in 2006 and 2008. In 2013 the NUbots missed out on a place in the quarter finals by only one goal. We hope to improve on this in 2014.

The NUbots joined the Kid-size league in 2012 with the DARwIn-OP robots, and ported the SPL NUbots codebase to the new platform. The NUbots retain a robust and fast vision and localisation system from their time in the SPL, and demonstrated the ability to draw on improvements in other leagues by porting the B-human NAO walk to the DARwIn-OP for 2012-2013. In 2013 the NUbots presented a curiosity based reinforcement learning approach to gaze planning at Robocup, which was used for all games during the competition. 


\section{Background of the NUbots Team Members}
\begin{itemize}
\item \emph{Josiah Walker} is studying for a Doctorate of Philosophy in Computer Science in Reinforcement Learning and Robotics. He works on robot behaviour and machine learning for various NUbot systems. He is the NUbots team leader for 2013-2014.

\item \emph{Brendan Annable} is a 3rd year undergraduate student studying Software Engineering currently working on network infrastructure as well as a browser-based debugging environment.

\item \emph{Alex Biddulph} is a 4th year undergraduate student studying Computer Engineering and Computer Science. His third year project is to improve the vision system and develop an alternative controller platform for the Darwin.

\item \emph{A/Prof. Stephan Chalup} is the head of the Newcastle Robotics
Laboratory. He is an Associate Professor in Computer Science and Software Engineering.
He is one of the initiators of the University of Newcastle's
RoboCup activities since 2001. His research area is machine learning
and anthropocentric robotics.

\item \emph{Andrew Dabson} is a fourth year undergraduate Engineering student (Software / Mechatronics), with a primary interest in bipedal locomotion and balance.

\item \emph{Jake Fountain} is an honours year undergraduate student studying computer science, with undergraduate degrees in mathematics and science. His main interests lie in machine learning and artificial intelligence.

\item \emph{Trent Houliston} is studying for a Doctorate of Philosophy in Software Engineering in Memory in Artificial Intelligence. He designed and implemented the new architecture for the robots, and assists in optimising many of the components.

\item \emph{Jordan Johnson} is a fourth year undergraduate student studying Computer Engineering and Computer Science. He is working on developing an audio communication system and a new electronics platform for the Darwin.

\item \emph{Taylor Johnson} is a fourth year undergraduate student studying a combined degree in Computer Engineering and Computer Science. He is working on improving vision and a new electronics platform for the Darwin.

\item \emph{Dr. Robert King} is a Lecturer in Statistics at the University of Newcastle. His research focus is on flexibly-shaped distributions,
statistical computing and Bayesian knowledge updating. He joined the
NUbots in 2004 and has developed a special interest in the RoboCup rules and refereeing.

\item \emph{Dr. Alexandre Mendes} is deputy head of the Newcastle Robotics Lab. He is a Senior Lecturer in Computer Science and Software Engineering. He joined the group in September 2011 and his research areas are algorithms and optimisation.

\item \emph{Mitchell Metcalfe} is a fourth year undergraduate student studying combined degree in Mathematics and Computer Science. He is working on the NUBots' localisation system, and maintains the support tools for the NUBots' compiler environment.

\item \emph{Peter Turner} is technical staff in the School of Electrical Engineering and Computer Science. Peter provides hardware support and assists the team with physical robot design upgrades. 

\item \emph{Anita Sugo} is a second year undergraduate student studying a combined degree in Mathematics and Science. She is interested in the mathematics used in robotics.

\end{itemize}

We also acknowledge the input of colleagues from the Newcastle Robotics Laboratory, team members of previous years
and the Interdisciplinary Machine Learning Research Group (IMLRG) in
Newcastle, Australia. Details are linked to the relevant webpages at
\texttt{www.robots.newcastle.edu.au}. 

\section{Hardware and Software Overview}
The NUbots use the DARwIn-OP robot with foot sensors. The team has five of these robots that are of the standard design with the exception of a slightly reduced foot size. The team also hopes to field modified DARwIn-OP consisting of a full HD camera, an ODroid-XU computer and an updated motor communications board as a part of a student project. 


The NUbots team's major research focus is on using machine learning methods within the software systems of the robot to achieve increased performance and autonomy~\cite{ChalupEtAlSMC2007}. The current NUbots software source is available from \cite{nubotsGit} and is covered under the GPL. This includes associated toolkits for building and deploying the software. Our software is designed to work on multiple robotic platforms, and all of the individual modules have been designed to be easily used in other systems. The sensors and actuators are accessed using a standard format, regardless of the robot running the software~\cite{Kulk2011c}. 

This year marks a major change in the software architecture of the NUBots, with improved software modularity coming from a cutting edge event based message passing system. The NUBots' architecture now has strong parallels with ROS (although without the associated overheads) and interoperability with ROS modules is planned. The NUBots software is designed to allow new teams and team members to easily understand and innovate on existing code. We plan to provide a full code release post Robocup 2014.



\section{Acknowledgement of Use of Code}
The NUbots DARwIn-OP robots use a walk engine ported from the 2013 Team Darwin code release. We acknowledge the source of this code. The NUBots have ported this code to C++ and restructured the logic, and also made changes including improving the inverse kinematic calculations. 

\section{Enhancements since RoboCup 2013}
Since Robocup 2013, improvements have been made in the area of software architecture and hardware. As the league moves to more realistic game conditions, we are trialling a hardware control platform composed of the following:

In response to the increase in field size, the Logitech C905 is being replaced with a Logitech C920. The new camera will provide a 1080p image, allowing the robots to detect and classify objects at greater distances.

The Main Controller (CompuLab fit-PC2i) is being replaced by an ODroid-XU to provide the extra processing power that is required for processing the images from the Logitech C920.

A prototype hardware replacement for the ROBOTIS CM730 Sub Controller, named the TAJ3850, is 
being developed as part of a Computer Engineering Third Year Project. The TAJ3850 provides power to all system components, a battery monitoring system featuring a low-voltage alarm and an extremely low-voltage automatic cut-off, an improved six-axis motion processing unit, a temperature sensor, and five dedicated motor buses. The TAJ3850 also exposes peripheral connections from the ODroid-XU to the back panel of the robot (HDMI, USB3, LAN, Audio output), monitors the three back-panel control buttons, and controls the status LEDs on the back panel and in the head.


The remaining robots will have the standard DARwIn-OP configuration of a Logitech C905, fit-PC2i and CM730.

\section{Research Areas}

\noindent\textbf{Robot Vision:} Vision is one of the major research areas associated with the Newcastle Robotics Laboratory. Several subtopics have been investigated including object recognition, horizon determination, edge detection, model fitting and colour classification using ellipse fitting, convex optimisation and kernel machines. Recent work has resulted in a fully-autonomous method of colour look-up table generation using k-means clustering and support vector machines, as well as evaluation of colour spaces for unsupervised learning and occluded feature detection. Publications are available e.g. from~\cite{budden2012colour,budden2012ball,henderson_2007,nickin_2007,NUBOT2006,Henderson2008,flannery2013ransac,budden2013salient}.
\\

\noindent\textbf{Localisation and Kalman Filters:} Research on the
topic of localisation focused on Bayesian approaches to robot
localisation including Kalman Filter and particle filter
based methods. We are interested in 
modifications of the Kalman Filter to handle non-ideal information
from vision, incorporate increased information from multiple agents, 
and effectively utilise non-unique objects.
\\

\noindent\textbf{Development of the Robot Bear:} In a collaborative
effort with the company Tribotix and colleagues in design a
bear-like robot (called Hykim) was developed~\cite{ChalupEtAl2006}. The idea was to have a modular open platform using high quality Dynamixel servos. 
\\

\noindent\textbf{Biped Robot Locomotion:} The improvement of walking speed and stability has been investigated by the NUbots for several years and on different platforms: On the AIBO robot we achieved one of the fastest walks at that time by walk parameter evolution \cite{QuinlanEtAlACRA2003,ChalupEtAlSMC2007}. On the Nao robot we improved existing walk engines by modifying the joint stiffnesses, or controller gains, \cite{Kulk2008,Kulk2010,Kulk2010a}. The stiffnesses were selected through an iterative process to maximise the cost of transport. We investigated the application of Support Vector Machines and Neural Networks to proprioception data for sensing perturbations during pseudo quiet stance. Walk improvements have been primarily done via optimisation techniques \cite{Kulk2011a},  
with recent improvements to our framework for online optimisation of bipedal humanoid locomotion. 
The use of spiking neural networks has been trialled in simulation~\cite{WiklendtChalup2008}. Prior to RoboCup 2012 the walk engine developed by the leading SPL team BHuman~\cite{BHumanWalk2010} was ported to the DARwIn-OP platform, and a variety of optimisation techniques were developed and successfully applied to improve walking speed and stability of the DARwIn-OP`\cite{budden2013probabilistic}. Similar optimisations are in process for the Team Darwin 2013 walk.\\

\noindent\textbf{Reinforcement Learning, Affective Computing and Robot Emotions:} We investigate the feasibility of reinforcement learning or neurodynamic programming for applications such as motor control and music composition. Concepts for affective computing are developed in multidisciplinary projects in collaboration with the areas of architecture and cognitive science. The concept of emotion is important for selective memory formation and action weighting and continues to gain importance in the robotics community, including within robotic soccer. A number of projects in the Newcastle Robots Laboratory already address this topic~\cite{Pareidolia2010,HongEtAl2013a,HongEtAl2013b,WongEtAl2013}. 
\\

\noindent\textbf{Gaze analysis and head movement behavioural learning:} We investigated methods for human and robot pedestrian gaze analysis in~\cite{JalalianEtAl_CAADRIA2011,WongEtAl2012} as well as space perception, way finding and the detection and analysis of salient regions~\cite{BhatiaChalup2013,BhatiaEtAl2013,BhatiaChalupOstwald2013}. Recently we  applied motivated reinforcement learning techniques to optimising head movement behaviour, providing a robust algorithm by which a robot learns to choose landmarks to localise efficiently during a soccer game~\cite{FountainEtAl2014}. This algorithm was used in competition at Robocup 2013.
\\

\noindent\textbf{Manifold Learning:} In several projects we
investigate the application of non-linear dimensionality reduction
methods in order to achieve more understanding of, and more precise 
and efficient processing of, high-dimensional visual and acoustic data.
\cite{ChalupEtAl2007b,WongChalup_WCCI_2008,WongEtAl2012}.
\\

\noindent\textbf{Other new projects:} Much work has been focused on the underlying software architecture and external utilities to enable flexibility and extensibility for future research. Projects undertaken include improving the configurability of the software system via real-time configuration updates, development of a web-based online visualisation and debugging utility \cite{AnnableEtAl2014} and the application of software architectural principles to create a multithreaded event-based system with almost no run-time overhead. Some of this work is currently still in progress as part of the 2013/2014 robotics lab summer projects by new undergraduate students who have joined the team.

\section{Related Research Concentrations}

The \emph{Interdisciplinary Machine Learning Research Group (IMLRG)} investigates different aspects of machine learning and data mining in theory, experiments and applications. Particular emphasis is put on interdisciplinary projects. The IMLRG's research areas include: Dimensionality reduction, vision processing, robotics control and learning, neurocomputing, evolutionary computation, optimisation, reinforcement learning, and kernel methods.

\noindent Links to publications can be found at the NUbots' webpage
\begin{center}
\texttt{http://robots.newcastle.edu.au/}
\end{center}

\bibliographystyle{plain}
\bibliography{nubots}

\begin{thebibliography}{10}

\bibitem{AnnableEtAl2014}
Brendan Annable, David Budden, and Alexandre Mendes.
\newblock Nubugger: A visual real-time robot debugging system.
\newblock In {\em RoboCup 2013: Robot Soccer World Cup XVII}, Lecture Notes in
  Artificial Intelligence (LNAI). Springer, 2014.
\newblock accepted 8.5.2013.

\bibitem{BhatiaEtAl2013}
Shashank Bhatia and Stephan~K. Chalup.
\newblock A model of heteroassociative memory: Deciphering surprising features
  and locations.
\newblock In Mary~L. Maher, Tony Veale, Rob Saunders, and Oliver Bown, editors,
  {\em Proceedings of the Fourth International Conference on Computational
  Creativity (ICCC 2013)}, pages 139--146, Sydney, Australia, June 2013.

\bibitem{BhatiaChalup2013}
Shashank Bhatia and Stephan~K. Chalup.
\newblock Segmenting salient objects in 3d point clouds of indoor scenes using
  geodesic distances.
\newblock {\em Journal of Signal and Information Processing}, 4(3B):102--108,
  2013.

\bibitem{BhatiaChalupOstwald2013}
Shashank Bhatia, Stephan~K. Chalup, and Michael~J. Ostwald.
\newblock Wayfinding: a method for the empirical evaluation of structural
  saliency using 3d isovists.
\newblock {\em Architectural Science Review}, 56(3):220--231, 2013.

\bibitem{budden2012colour}
D.~Budden, S.~Fenn, A.~Mendes, and S.~Chalup.
\newblock Evaluation of colour models for computer vision using cluster
  validation techniques.
\newblock In {\em RoboCup 2012: Robot Soccer World Cup XVI}, Lecture Notes in
  Computer Science. Springer, 2013.

\bibitem{budden2012ball}
D.~Budden, S.~Fenn, J.~Walker, and A.~Mendes.
\newblock A novel approach to ball detection for humanoid robot soccer.
\newblock In {\em Advances in Artificial Intelligence (LNAI 7691)}. Springer,
  2012.

\bibitem{budden2013salient}
D.~Budden and A.~Mendes.
\newblock Unsupervised recognition of salient colour for real-time image
  processing.
\newblock In {\em RoboCup 2013: Robot Soccer World Cup XVII}. Springer, 2013.

\bibitem{budden2013probabilistic}
D.~Budden, J.~Walker, M.~Flannery, and A.~Mendes.
\newblock Probabilistic gradient ascent with applications to bipedal robot
  locomotion.
\newblock In {\em Australasian Conference on Robotics and Automation (ACRA)},
  2013.

\bibitem{ChalupEtAl2006}
S.~K. Chalup, M.~Dickinson, R.~Fisher, R.~H. Middleton, M.~J. Quinlan, and
  P.~Turner.
\newblock Proposal of a kit-style robot as the new standard platform for the
  four-legged league.
\newblock In {\em Australasian Conference on Robotics and Automation (ACRA)
  2006}, 2006.

\bibitem{ChalupEtAl2007b}
Stephan~K. Chalup, Riley Clement, Joshua Marshall, Chris Tucker, and Michael~J.
  Ostwald.
\newblock Representations of streetscape perceptions through manifold learning
  in the space of hough arrays.
\newblock In {\em 2007 IEEE Symposium on Artificial Life}, 2007.

\bibitem{Pareidolia2010}
Stephan~K. Chalup, Kenny Hong, and Michael~J. Ostwald.
\newblock Simulating pareidolia of faces for architectural image analysis.
\newblock {\em International Journal of Computer Information Systems and
  Industrial Management Applications (IJCISIM)}, 2:262--278, 2010.

\bibitem{ChalupEtAlSMC2007}
Stephan~K. Chalup, Craig~L. Murch, and Michael~J. Quinlan.
\newblock Machine learning with aibo robots in the four legged league of
  robocup.
\newblock {\em IEEE Transactions on Systems, Man, and Cybernetics---Part C},
  37(3):297--310, May 2007.

\bibitem{ChalupOstwald2009}
Stephan~K. Chalup and Michael~J. Ostwald.
\newblock Anthropocentric biocybernetic computing for analysing the
  architectural design of house facades and cityscapes.
\newblock {\em Design Principles and Practices: An International Journal},
  3(5):65--80, 2009.

\bibitem{flannery2013ransac}
Madison Flannery, Shannon Fenn, and David Budden.
\newblock Ransac: Identification of higher-order geometric features and
  applications in humanoid robot soccer.
\newblock {\em arXiv preprint arXiv:1310.5781}, 2013.

\bibitem{FountainEtAl2014}
Jake Fountain, Josiah Walker, David Budden, Alexandre Mendes, and Stephan~K.
  Chalup.
\newblock Motivated reinforcement learning for improved head actuation of
  humanoid robots.
\newblock In {\em RoboCup 2013: Robot Soccer World Cup XVII}, Lecture Notes in
  Artificial Intelligence (LNAI). Springer, 2014.
\newblock accepted 8.5.2013.

\bibitem{BHumanWalk2010}
Colin Graf and Thomas R{\"o}fer.
\newblock A closed-loop 3d-lipm gait for the robocup standard platform league
  humanoid.
\newblock In Enrico Pagello, Changjiu Zhou, Sven Behnke, Emanuele Menegatti,
  Thomas R{\"o}fer, and Peter Stone, editors, {\em Proceedings of the Fifth
  Workshop on Humanoid Soccer Robots in conjunction with the 2010 IEEE-RAS
  International Conference on Humanoid Robots}, Nashville, TN, USA, 2010.

\bibitem{Henderson2008}
N.~Henderson, R.~King, , and S.K. Chalup.
\newblock An automated colour calibration system using multivariate gaussian
  mixtures to segment hsi colour space.
\newblock In {\em Proc. of the 2008 Australasian Conference on Robotics and
  Automation}, 2008.

\bibitem{henderson_2007}
N.~Henderson, R.~King, and R.~H. Middleton.
\newblock An application of gaussian mixtures: Colour segmenting for the four
  legged league using hsi colour space.
\newblock In {\em RoboCup Symposium, Atlanta, July 2007}, Lecture Notes in
  Computer Science, 2007.

\bibitem{HongEtAl2013a}
Kenny Hong, Stephan Chalup, and Robert King.
\newblock A component based approach for classifying the seven universal facial
  expressions of emotion.
\newblock In {\em IEEE Symposium on Computational Intelligence for Creativity
  and Affective Computing 2013}. IEEE, 2013.

\bibitem{HongEtAl2013b}
Kenny Hong, Stephan~K. Chalup, and Robert King.
\newblock Scene perception using pareidolia of faces and expressions of
  emotion.
\newblock In {\em IEEE Symposium on Computational Intelligence for Creativity
  and Affective Computing 2013}. IEEE, 2013.

\bibitem{JalalianEtAl_CAADRIA2011}
Arash Jalalian, Stephan~K. Chalup, and Michael~J. Ostwald.
\newblock Agent-agent interaction as a component of agent-environment
  interaction in the modelling and analysis of pedestrian visual behaviour.
\newblock In {\em CAADRIA 2011. Circuit Bending, Breaking and Mending. The 16th
  International Conference of the Association for Computer-Aided Architectural
  Design Research in Asia}, 2011.

\bibitem{Kulk2010a}
J.~Kulk and J.~Welsh.
\newblock Perturbation sensing using proprioception for humanoid robots.
\newblock In {\em Proceedings of the IEEE Conference on Robotics and
  Automation}, 2010.

\bibitem{Kulk2008}
J.A. Kulk and J.S. Welsh.
\newblock A low power walk for the nao robot.
\newblock In {\em Proc. of the 2008 Australasian Conference on Robotics and
  Automation (ACRA'2008)}, 2008.

\bibitem{Kulk2010}
J.A. Kulk and J.S. Welsh.
\newblock Autonomous optimisation of joint stiffnesses over the entire gait
  cycle for the nao robot.
\newblock In {\em Proceedings of the 2010 International Symposium on Robotics
  and Intelligent Sensors.}, 2010.

\bibitem{Kulk2011a}
Jason Kulk and James Welsh.
\newblock Evaluation of walk optimisation techniques for the nao robot.
\newblock In {\em IEEE-RAS International Conference on Humanoid Robots}, 2011.

\bibitem{Kulk2011c}
Jason Kulk and James Welsh.
\newblock A nuplatform for software on articulated mobile robots.
\newblock In {\em 1st International ISoLA Workshop on Software Aspects of
  Robotic Systems}, 2011.

\bibitem{nubotsGit}
M.~Metcalfe, J.~Fountain, A.~Sugo, T.~Houliston, A.~Buddulph, A.~Dabson,
  T.~Johnson, J.~Johnson, B.~Annable, , S.~Nicklin, S.~Fenn, D.~Budden,
  J.~Walker, and J.~Reitveld.
\newblock Nubots robocup code repository.
\newblock \url{https://github.com/nubots/NUClearPort}, January 2014.

\bibitem{nickin_2007}
S.P Nicklin, R.~Fisher, and R.H. Middleton.
\newblock Rolling shutter image compensation.
\newblock In {\em Robocup Symposium 2006}, 2007.

\bibitem{QuinlanEtAlACRA2003}
M.~J. Quinlan, S.~K. Chalup, and R.~H. Middleton.
\newblock Techniques for improving vision and locomotion on the aibo robot.
\newblock In {\em Australian Conference on Robotics and Automation
  (ACRA'2003)}. ARAA (on-line), 2003.

\bibitem{NUBOT2006}
M.J. Quinlan, S.P. Nicklin, N.~Henderson, Fisher R., F.~Knorn, S.K. Chalup,
  R.H. Middleton, and R.~King.
\newblock The 2006 nubots team report.
\newblock Technical report, School of Electrical Engineering and Computer
  Science, The University of Newcastle, Australia, 2006.

\bibitem{WiklendtChalup2008}
L.~Wiklendt, S.~K. Chalup, and M.~M. Seron.
\newblock Simulated 3d biped walking with and evolution-strategy tuned spiking
  neural network.
\newblock {\em Neural Network World}, 19:235--246, 2009.

\bibitem{WongChalup_WCCI_2008}
Aaron~S.~W. Wong and Stephan~K. Chalup.
\newblock Towards visualisation of sound-scapes through dimensionality
  reduction.
\newblock In {\em 2008 International Joint Conference on Neural Networks (IJCNN
  2008)}, pages 2833--2840. IEEE, 2008.

\bibitem{WongEtAl2012}
Aaron~S.~W. Wong, Stephan~K. Chalup, Shashank Bhatia, Arash Jalalian, Jason
  Kulk, Steven Nicklin, and Michael~J. Ostwald.
\newblock Visual gaze analysis of robotic pedestrians moving in urban space.
\newblock {\em Architectural Science Review}, 55(3):213--223, 2012.

\bibitem{WongEtAl2013}
Aaron~S.W. Wong, Kenny Hong, Steven Nicklin, Stephan~K. Chalup, and Peter
  Walla.
\newblock Robot emotions generated and modulated by visual features of the
  environment.
\newblock In {\em IEEE Symposium on Computational Intelligence for Creativity
  and Affective Computing 2013}. IEEE, 2013.

\end{thebibliography}

\end{document}